# On Model Extrapolation in Marginal Shapley Values

Ilya Rozenfeld

Capital One Financial
12/02/2024

## Abstract

As the use of complex machine learning models continues to grow, so does the need for reliable explainability methods. One of the most popular methods for model explainability is based on Shapley values.  There are two most commonly used approaches to calculating Shapley values which produce different results when features are correlated, conditional and marginal. In our previous work, it was demonstrated that the conditional approach is fundamentally flawed due to implicit assumptions of causality. However, it is a well-known fact that marginal approach to calculating Shapley values leads to model extrapolation where it might not be well defined.

In this paper we explore the impacts of model extrapolation on Shapley values in the case of a simple linear spline model. Furthermore, we propose an approach which while using marginal averaging avoids model extrapolation and with addition of causal information replicates causal Shapley values. Finally, we demonstrate our method on the real data example.

## 1   Introduction

In recent years, the rapid advancement of machine learning (ML) and artificial intelligence (AI) has revolutionized various domains, from healthcare to finance, enabling the creation of highly accurate predictive models. However, as these models grow in complexity, their inner workings become increasingly opaque, leading to a significant challenge of interpreting model predictions. Understanding how models arrive at their predictions is crucial for building trust, ensuring fairness, and facilitating decision-making in high-stakes environments.

One of the most promising and popular approaches to model interpretability is based on Shapley values, a concept originally developed in cooperative game theory [1] which we briefly describe here. Given a set of players $\mathcal{M} = \{1, \ldots, M\}$ and a measure of payoff that can be achieved by cooperating players in coalition $S$ as value function $v(S)$ defined for all possible $S \subseteq \mathcal{M}$, the goal is to determine $j^{th}$ player contribution $\phi_j(v)$ to the total payoff $v(\mathcal{M})$.

To make attributions fair, four axioms are usually imposed:

1. **Efficiency**: Total payout is fully distributed across players

$$v(\mathcal{M}) = \sum_{j \in \mathcal{M}} \phi_j(v)$$

2. **Dummy:** A player that doesn't contribute to any coalition receives zero attribution
$$v(S \cup \{j\}) = v(S), \forall S \implies \phi_j(v) = 0$$

3. **Symmetry**: Two players that contribute equally to every coalition get the same attributions
$$v(S \cup \{j\}) = v(S \cup \{i\}), \forall S \implies \phi_j(v) = \phi_i(v)$$

4. **Additivity**: Sum of player's attributions from two different games is equal to attribution from combined game
$$\phi_i(v_1 + v_2) = \phi_i(v_1) + \phi_i(v_2)$$

With the addition of the axioms the attribution problem has unique solution

$$\phi_j(v) = \sum_{S \subseteq \mathcal{M} \setminus \{j\}} \frac{|S|! \, (M - |S| - 1)!}{M!} \big(v(S \cup \{j\}) - v(S)\big) \tag{2.1}$$

where $|S|$ denotes the number of players and $\mathcal{M} \setminus \{j\}$ a set of all players excluding $j^{th}$ one. The formula for $\phi_j(v)$ can be viewed as the weighted sum of the marginal contributions $v(S \cup \{j\}) - v(S)$ of $j^{th}$ player. The quantities $\phi_j(v)$ are referred to as Shapley Values.

The application of the Shapley game-theoretic framework to ML and AI models is referred to as **SH**apley **A**dditive ex**P**lanation (SHAP) [2]. Suppose we have an ML or AI model $f(X)$ with a set of features $X = \{X_1, \ldots, X_M\}$. Additionally, suppose we want to explain model's prediction at some sample $X = x^*$. To apply cooperative game theory to model explanation we can view prediction $f(x^*)$ as total payout and features $X$ as players with complimentary subsets $X_S$ and $X_{\bar{S}}$ (i.e. $X = \{X_S, X_{\bar{S}}\}$) being as in-coalition and out-of-coalition, respectively.

Defining the last necessary component of the game, the value function $v(S)$, is not as straightforward. The reason is that we need to be able to evaluate the model on the subset of features $x_S^*$ while it requires the full set $x^*$. Among different approaches [3], the two by far most popular are conditional and marginal averages over out-of-coalition features. The value function with conditional average is defined as

$$v(S) = E[f(X)|X_S = x_S^*] \tag{2.2}$$

and with marginal as

$$v(S) = E[f(\{x_S^*, X_{\bar{S}}\})] \tag{2.3}$$

Then the prediction $f(x^*)$ can be decomposed into additive explanations as

$$f(x^*) = \phi_0 + \sum_{j=1}^{M} \phi_j(x^*) \tag{2.4}$$

where $\phi_0 = E[f(X)]$.

While both, marginal and conditional averaging, produce the same results when features are independent, they differ for correlated features. This raises an issue as to which method is the correct one. In [4] marginal averaging was justified based on causal arguments by showing that it corresponds to Pearls do-operator [5]. Additionally, in [6] it was further shown that the conditional averaging is fundamentally flawed as it implicitly makes causal assumptions based on the presence of correlations. For these reasons both references recommend using marginal averaging. However, the issue with marginal averaging as was pointed out in numerous works (see [7] for example) is that it forces the model to extrapolate into feature regions with sparse or no data.

In this paper we attempt to explore and address some of the issues connected to the model extrapolation in marginal averaging. First, we demonstrate the problems that are caused by the model extrapolation using a simple linear spline model (Section 3). Second, in Section 4 we propose an approach that while using marginal averaging avoids model extrapolation. Lastly, the proposed approach is applied to real data in Section 5.

## 2  Shapley Values for Linear Spline Model

To explore the impact of the extrapolation on Shapley Values let us consider the model

$$f(\mathbf{X}) = \beta_0 + \beta_1 X_1 + \beta_{12} X_1 X_2 \tag{3.1}$$

where $X_1 \sim N(0,1)$ and $X_2 = I(X_1 > 0)$ i.e. $X_2 = 1$ when $X_1$ is positive and 0 otherwise. Since likelihood of $X_1 > 0$ is 0.5, $X_2$ is Bernoulli random variable with probability parameter equal to 0.5. This is an equation for linear spline with two different slopes in regions $X_1 > 0$ and $X_1 \leq 0$ as shown in Figure 1 by solid blue line. The features $X_1$ and $X_2$ are correlated with value $\sqrt{2/\pi} \approx 0.8$. The observations in feature regions $(X_1 < 0, X_2 = 1)$ and $(X_1 > 0, X_2 = 0)$ are impossible. However, computations of marginal Shapley values would force a model to extrapolate into those regions due to correlation between the features.

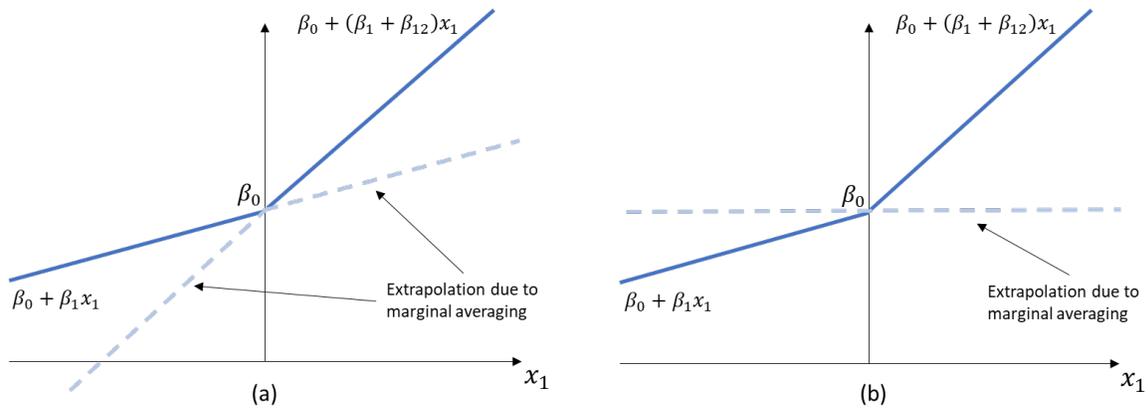

Figure 1: Linear spline model with linear (a) and constant (b) extrapolations due to marginal averaging.

### 2.1  Linear Extrapolation

If a parametric model specified by Eq. (3.1) was fitted to the data, the extrapolation would be linear as shown by dashed lines in Figure 1a. The value functions can be calculated in straightforward manner as

$$v(\emptyset) = \beta_0 + \beta_{12}\gamma$$

$$v(\{1\}) = \beta_0 + \beta_1 x_1^* + \frac{\beta_{12} x_1^*}{2}$$

$$v(\{2\}) = \beta_0$$

$$v(\{1,2\}) = \beta_0 + \beta_1 x_1^* + \beta_{12} x_1^* x_2^*$$

(3.2)

where $\gamma = E[X_1 X_2] = \sqrt{1/2\pi}$. Recalling the Shapley values formulas for a model with two features

$$\phi_0 = v(\emptyset)$$

$$\phi_1 = \frac{1}{2}\left(v(\{1,2\}) - v(\{2\})\right) + \frac{1}{2}\left(v(\{1\}) - v(\emptyset)\right)$$

$$\phi_2 = \frac{1}{2}\left(v(\{1,2\}) - v(\{1\})\right) + \frac{1}{2}\left(v(\{2\}) - v(\emptyset)\right)$$

(3.3)

and substituting value functions from Eq. (3.2) we arrive at Shapley values for the linear spline model with linear extrapolation as shown in Table 1.

Table 1: Shapley Values for linear spline model with linear extrapolation

|  | $x_1^* \leq 0, \; x_2^* = 0$ | $x_1^* > 0, \; x_2^* = 1$ |
|---|---|---|
| $\phi_0$ | $\beta_0 + \beta_{12}\gamma$ | $\beta_0 + \beta_{12}\gamma$ |
| $\phi_1$ | $\left(\beta_1 + \frac{\beta_{12}}{4}\right)x_1^* - \frac{\gamma\beta_{12}}{2}$ | $\left(\beta_1 + \frac{3\beta_{12}}{4}\right)x_1^* - \frac{\gamma\beta_{12}}{2}$ |
| $\phi_2$ | $-\frac{\beta_{12}}{4}x_1^* - \frac{\gamma\beta_{12}}{2}$ | $\frac{\beta_{12}}{4}x_1^* - \frac{\gamma\beta_{12}}{2}$ |

The presence of the terms proportional to $\beta_{12} x_1^*$ for $x_1^* \leq 0$ in both $\phi_1$ and $\phi_2$ is unintuitive since the parameter $\beta_{12}$ is not a part of the model in this region (Figure 1). For $x_1^* > 0$, on the other hand, the presence of the terms proportional to $\beta_{12} x_1^*$ is justifiable, but their weights (3/4 for $\phi_1$ and 1/4 for $\phi_2$) seem to be arbitrary. In fact, one could question the presence of this term in $\phi_2$ altogether since it is not clear why the impact of $X_2$ should increase as $|x_1^*|$ increases. These biases can become quite large if either or both $\beta_{12}$ and $|x_1^*|$ are large.

## 2.2 Constant Extrapolation

Next, let us assume that a decision tree model was fitted to the data with the first split on $x_2$ at 0 and the rest of the splits are on $x_1$ only. This extrapolation will be along horizontal lines as shown in Figure 1b by dashes. The calculations for this case are a bit more cumbersome than those for a linear case and are relegated to the Appendix A. The resulting Shapley Values for two feature regions are shown in Table 2.

Table 2: Shapley Values for linear spline model with constant extrapolation

|  | $x_1^* \leq 0, \quad x_2^* = 0$ | $x_1^* > 0, \quad x_2^* = 1$ |
|---|---|---|
| $\phi_0$ | $\beta_0 + \beta_{12}\gamma$ | $\beta_0 + \beta_{12}\gamma$ |
| $\phi_1$ | $\dfrac{3\beta_1 x_1^*}{4} + \dfrac{(\beta_1 - \beta_{12})\gamma}{2}$ | $\dfrac{3(\beta_1 + \beta_{12})x_1^*}{4} - \left(\beta_{12} + \dfrac{\beta_1}{2}\right)\gamma$ |
| $\phi_2$ | $\dfrac{\beta_1 x_1^*}{4} - \dfrac{(\beta_1 + \beta_{12})\gamma}{2}$ | $\dfrac{(\beta_1 + \beta_{12})x_1^*}{4} + \dfrac{\beta_1 \gamma}{2}$ |

Unlike for the case of linear extrapolation, there are no terms proportional to $\beta_{12}x_1^*$ in the region $x_1^* \leq 0$. Nevertheless, the presence of the terms proportional to $\beta_1 x_1^*$ in both $\phi_1$ and $\phi_2$ leads to similar problems as in case of liner extrapolation: seemingly arbitrary splitting of the terms between $\phi_1$ and $\phi_2$ and increasing impact of $X_2$ with increasing $|x_1^*|$.

Another issue is that if data has some randomness, a different similarly performing tree model could be fitted with splits on $x_2$ happening further down the line resulting in reduction in $\phi_2$ and increase in $\phi_1$. In fact, the model could be built without any $x_2$ splits at all resulting in $\phi_2 = 0$.

## 3 Stratified Approach to Calculating Shapley Values

As was shown in the previous section the model extrapolation in computing marginal Shapley values can lead to undesired effects such unintuitive terms and sensitivity to the type of the model. Continuing with an example of linear spline in Eq. (3.1) we will explore approaches that avoid extrapolations and produce reasonable Shapley Values.

Let us consider calculating Shapley values separately for feature regions $(X_1 > 0, X_2 = 1)$ and $(X_1 < 0, X_2 = 0)$ as if they were two separate models[1]. $X_1$ would have positive and negative half-normal distributions in those regions with means $2\gamma$ and -$2\gamma$, respectively. Since the models are linear in each region, the Shapley values are straight forward to calculate and are shown in Table 3.

---

[1] In practice there is no need for two separate models. Instead, an appropriate background data when calculating Shapley values can be used [13].

Table 3: Shapley values for two separate regions

|  | $x_1^* \leq 0, \quad x_2^* = 0$ | $x_1^* > 0, \quad x_2^* = 1$ |
|---|---|---|
| $\phi_0$ | $\beta_0 - 2\beta_1\gamma$ | $\beta_0 + 2(\beta_1 + \beta_{12})\gamma$ |
| $\phi_1$ | $\beta_1(x_1^* + 2\gamma)$ | $(\beta_1 + \beta_{12})(x_1^* - 2\gamma)$ |

While this approach addresses some of the shortcomings by producing the expressions for $\phi_1$ that look intuitive, it is unsatisfactory. First, there is no longer any information about $X_2$ contribution. Second, there are two different reference values, $\phi_0$, in each region which begs its own explanation. Basically, non-constancy of $\phi_0$ implies that it contains the effects that should be attributed to either $\phi_1$ or $\phi_2$. To obtain the usual Shapley values which would include $\phi_2$, portions of $\phi_0$ in each feature region would need to be redistributed back to $\phi_1$ and $\phi_2$ in such a way that new $\phi_0$ becomes constant.

Exactly how $\phi_0$ should be redistributed can be determined by considering causality. When the linear spline example was specified, there was no mention of a causal relationship between the two features. Let us now introduce this information into the problem. First, let us assume that the sign of $X_1$ completely determines the value $X_2$, i.e. $X_1 \to X_2$. In this case $X_2$ is redundant implying that $\phi_2$ should be 0, and portion of $\phi_0$ should be transferred to $\phi_1$ only. The resulting Shapley Values are shown in the upper half of Table 4. The remaining constant portion of $\phi_0$ is denoted by $\xi$, and there is a freedom in choosing its value. We will discuss some potential choices later in this section. In the remainder of the paper this approach will be referred to as "stratified".

Next, let us assume that $X_2$ determines the sign of $X_1$, i.e. $X_2 \to X_1$, determining $X_1$ only partially since it has no impact on magnitude of $X_1$. In this case, since the impact of $X_1$ is already captured in $\phi_1$, it makes sense to transfer a portion of $\phi_0$ to $\phi_2$ with resulting Shapley Values shown in the lower half of Table 4. The interpretation is that $X_2$ controls in what feature region an observation belongs.

The freedom to choose $\xi$ could allow us to answer different questions. Choosing $\xi$ as either value of $\phi_0$ in Table 3 would provide an answer to the question, "What distinguishes an observation in one region from the average prediction in the same or another region?". One could also choose $\xi$ as some representative observation in which case the question being answered would be "What distinguishes an observation in one region from representative observation in the same or another region?". Lastly, one could set $\xi$ to a total average ($\phi_0$ from Table 2) to answer a more typical question for Shapley values, "What distinguishes an observation from the mean prediction of the model?". However, some caution is necessary in selecting the values of $\xi$ as making them too large or small can significantly distort a feature's Shapley values.

Table 4: Stratified Shapley values for linear spline model

|  |  | $x_1^* \leq 0, \quad x_2^* = 0$ | $x_1^* > 0, \quad x_2^* = 1$ |
|---|---|---|---|
| $X_1 \to X_2$ | $\phi_0$ | $\xi$ | $\xi$ |
|  | $\phi_1$ | $\beta_0 + \beta_1 x_1^* - \xi$ | $\beta_0 + (\beta_1 + \beta_{12}) x_1^* - \xi$ |
|  | $\phi_2$ | 0 | 0 |
| $X_2 \to X_1$ | $\phi_0$ | $\xi$ | $\xi$ |
|  | $\phi_1$ | $\beta_1(x_1^* + 2\gamma)$ | $(\beta_1 + \beta_{12})(x_1^* - 2\gamma)$ |
|  | $\phi_2$ | $\beta_0 - 2\beta_1\gamma - \xi$ | $\beta_0 + 2(\beta_1 + \beta_{12})\gamma - \xi$ |

Another approach that avoids extrapolation is to use causal (or asymmetric) Shapley values from [8] and [9]. As shown in Appendix B, this approach leads to the same result as in Table 4 with $\xi = \beta_0 + \beta_{12}\gamma$. It is notable how very different uses of the same causal information led to the same result. The equivalence of the results shows that causal Shapley values can be computed without conditional averaging which is usually more complicated and computationally expensive [10]. Another consequence of the equivalence is that stratified Shapley values violate the Symmetry axiom.

## 4   Real Data Example

We apply stratified Shapley values method developed in previous section to the French motor third party liability claims frequency data studied in [11][2]. The dataset contains around 680 thousand observations with the following features:

1. IDpol: policy number (unique identifier)
2. ClaimNb: number of claims on the given policy
3. Exposure: total exposure in yearly units
4. Area: area code (categorical, ordinal)
5. VehPower: power of the car (categorical, ordinal)
6. VehAge: age of the car in years
7. DrivAge: age of the (most common) driver in years
8. BonusMalus: bonus-malus level between 50 and 230 (with reference level 100)

---
[2] The data is available at https://www.kaggle.com/datasets/floser/french-motor-claims-datasets-fremtpl2freq

9. VehBrand: car brand (categorical, nominal)
10. VehGas: diesel or regular fuel car (binary)
11. Density: density of inhabitants per km2 in the city of the living place of the driver
12. Region: regions in France (prior to 2016), these are illustrated in Figure 1 (categorical)

The first feature, IDpol, has no predictive power. The second feature is the target, and the third one will be used as an offset in the Poisson regression model (see Sec. 5.2.3 in [11]). This leaves us with 9 explanatory features.

To start, we explore correlation between the features in Figure 2. There are two pairs of features with high correlations, Area/Density and Driver's Age/Bonus Malus. All other correlations are relatively small. Given extremely high Area/Density correlation, only one of these features, Area, will be used in the model.

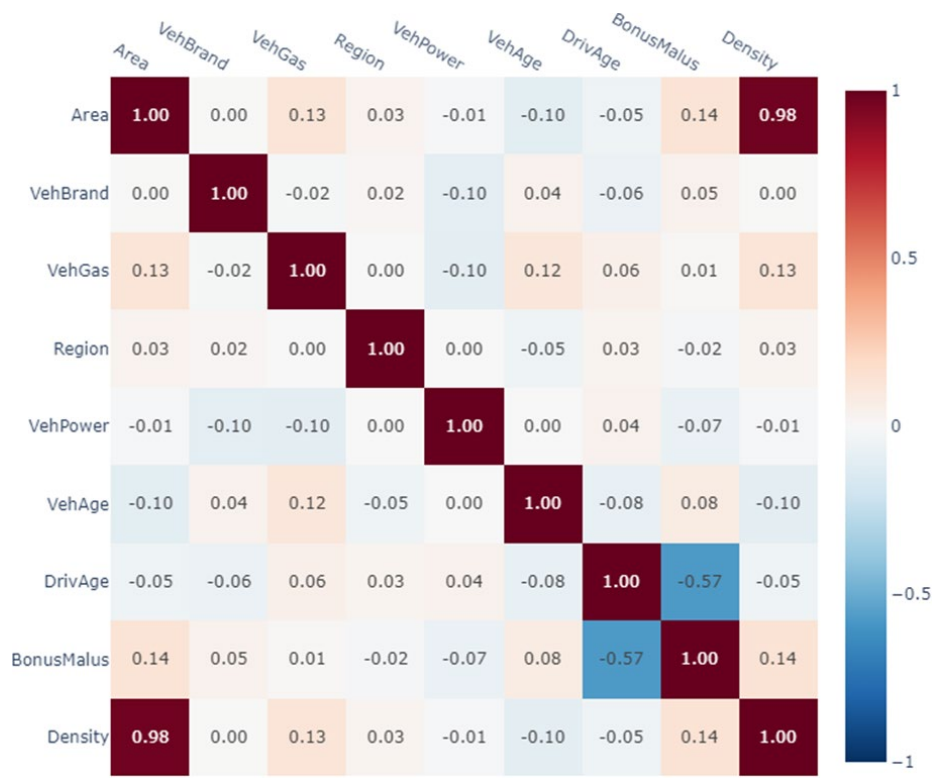

Figure 2: Spearman correlation matrix.

The dependency between Driver's Age and Bonus Malus is further explored in Figure 3. The blue bars show distributions of Bonus Malus values for each age with red lines denoting the medians. As expected from high correlation value between these two features, the plot shows strong dependency between the two features. Individuals older than 35 years, a majority of the population, tend to have Bonus Malus values close to 50. Very young people, aged between 18 and 23 years, on the other hand, tends to have score close to 100 as it takes some time to deviate from reference value. It is fairly rare for this group to have score close to 50 leading to sparse training data in this region[3].

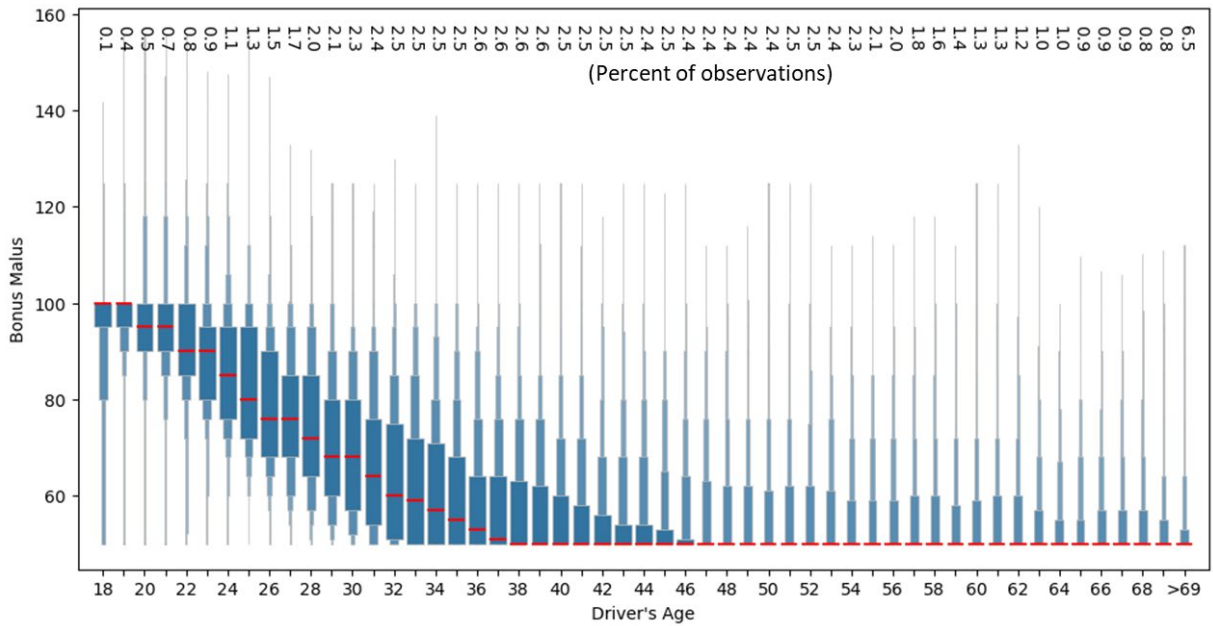

Figure 3: Dependency of Bonus Malus on driver's age.

Proceeding to the modeling, the data is split into training (80%) and test (20%) datasets. The training data is used to fit the Poisson regression model utilizing Histogram-Based Gradient Boosting (HGBT) method from scikit-learn (Section1.11.1.1 in [12]). The model's prediction we are interested in is the logarithm of the frequency of claims.

The testing data was used to compute Shapley values for 50 randomly selected observations for each age from 18 through 80 for a total of 3,150 observations[4]. We computed two types of Shapley values, the standard and stratified. For the computation of the latter, a random subsample of training data with the same age as that of observation was used as a background. In other words, a different background data was used for each age group.

The scatter plots of standard vs. stratified Shapley values are displayed in Figure 4 with the Bonus Malus feature shown separately in panel (a) and the rest of the features in panel (b). The closer the values from two methods to each other, the closer the points will lie to the reference 45º line (red dashes). As might be expected, due to strong correlation between driver's age and Bonus Malus, the stratified Shapley values for Bonus Malus differ significantly from those from the standard method. The line fitted to the scatter plot (dashed yellow line) has a slope that is significantly different from 1. Additionally, the differences are strongly age dependent with stratified Shapley values higher than standard ones for older drivers and lower for younger ones. The Shapley values from the two approaches for other features are similar as can be seen from Figure 4b where points are hugging the 45º reference line, and the fitted line has a slope of nearly 1.

---

[3] This observation was used as a rationale for using conditional Shapley values for this data in [14]
[4] We used Permutation explainer from SHAP package (https://github.com/shap/shap)

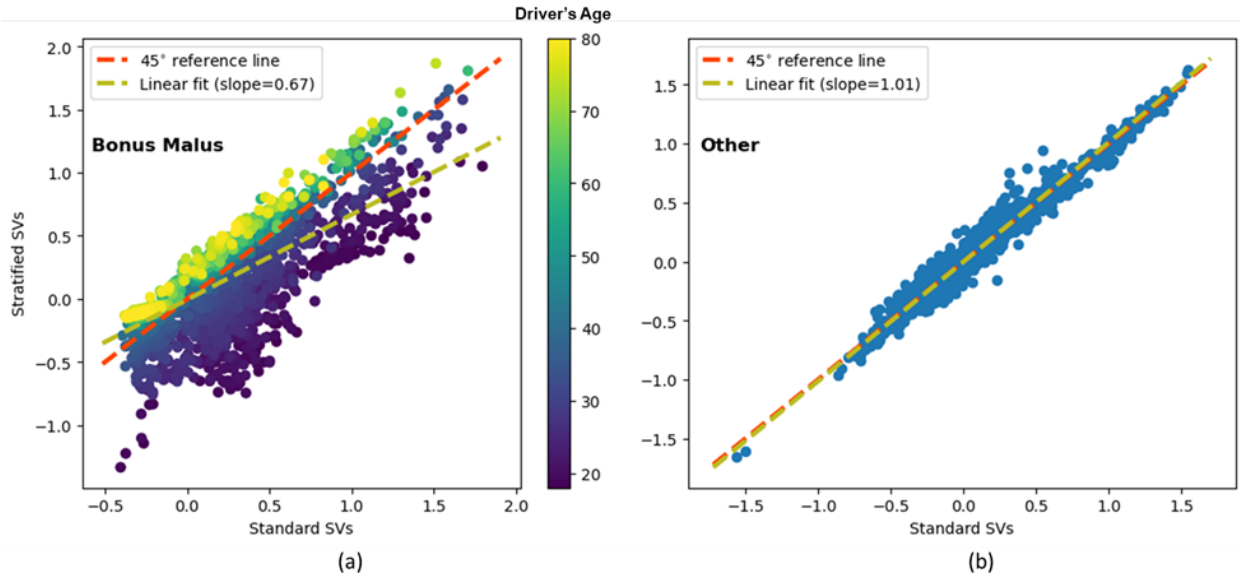

Figure 4: Comparison of standard and stratified Shapley values for Bonus Malus feature (a) and the rest of the features (b).

Note that Figure 4b doesn't include driver's age feature since as was discussed in Section 3, by calculating Shapley values for each age group separately, we no longer have information about its contribution to the prediction and also have non-constant age dependent reference model (blue curve in Figure 5). The reference values for individuals older than 27 years are close to the reference value from standard approach (red line in Figure 5) but differ significantly from that value for the younger population. Following the methodology proposed in Section 3, a non-constant portion of reference model $\phi_0$ should be transferred to the driver's age feature since it is obvious that the age is the cause of the Bonus Malus score. The resulting Shapley values with constant reference model chosen to be the same as for standard approach (red line in Figure 5) are shown on the left-hand side of Figure 6.

For comparison the Shapley values from the standard approach are shown on the right-hand side of Figure 6. As expected, the Shapley values for driver's age and Bonus Malus features have very different patterns for the two approaches. The driver's age contributes relatively little for the older individuals in the stratified approach. The contributions are a bit larger for the standard one. The contributions are relatively similar for very young drivers for both approaches. However, for the ages between 30 and 40, the standard approach produces a negative contribution not evident in stratified approach.

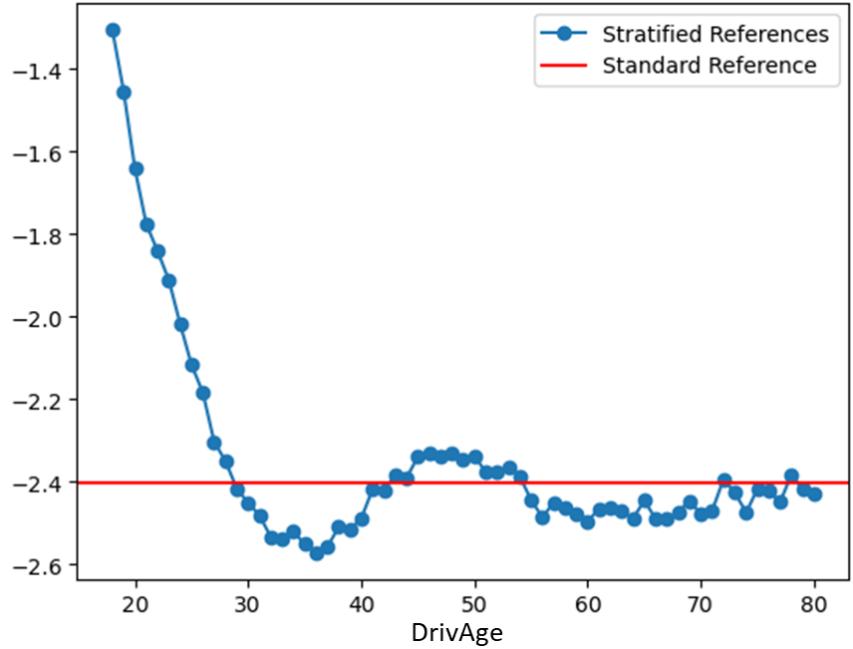

Figure 5: Reference models for stratified and standard approaches.

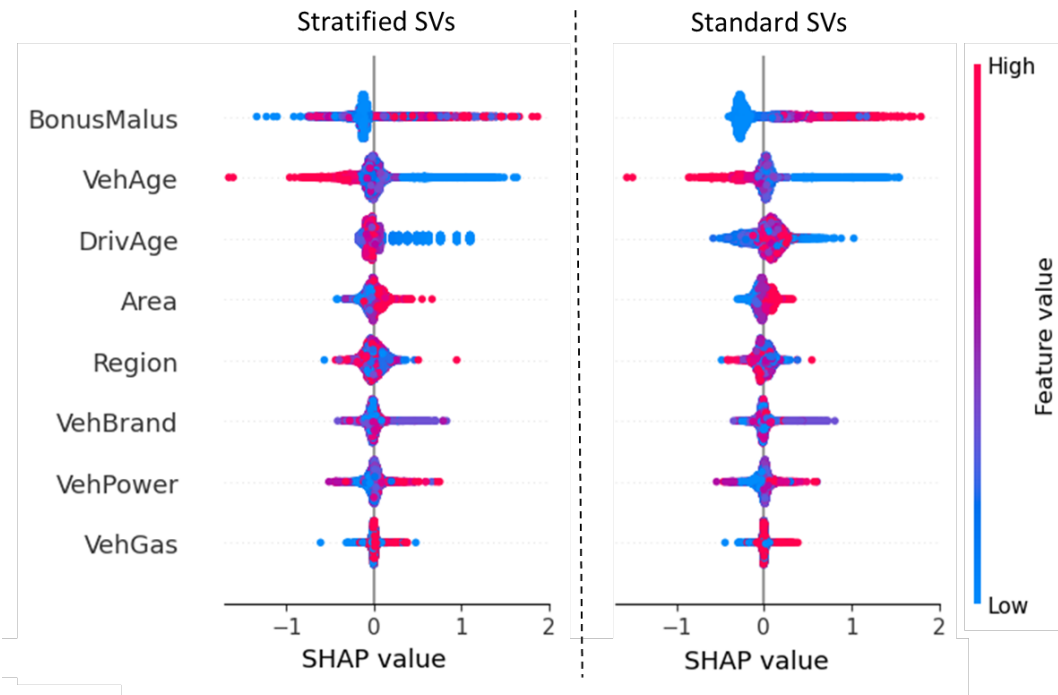

Figure 6: Stratified (with causal adjustment) and standard Shapley values. Both have the same reference value.

A more detailed comparison of Bonus Malus Shapley values from the two approaches is shown in Figure 7. The dependence of Bonus Malus contributions on driver's age is much stronger for stratified approach. In fact, for Bonus Malus values near 50, Shapley values from the standard approach show no age dependence at all. For higher Bonus Malus values, the dependence of Shapley values on age is inconsistent with lower values for older people for some Bonus Malus values (e.g. 100 and 120) and vice versa for others (e.g. 80). For the stratified approach on the other hand, not only the Shapley values are spread over a wider range, but they tend to be consistently higher for every value of Bonus Malus for older individuals implying strong interaction.

Note a few very low stratified Shapley values for very young individuals with Bonus Malus 50. A possible explanation could be that a very young person would need to be an extremely safe driver to achieve such a low score. Alternatively, this could be a signal that these observations are anomalous and further investigation is warranted. This insight wasn't evident from the standard Shapley values.

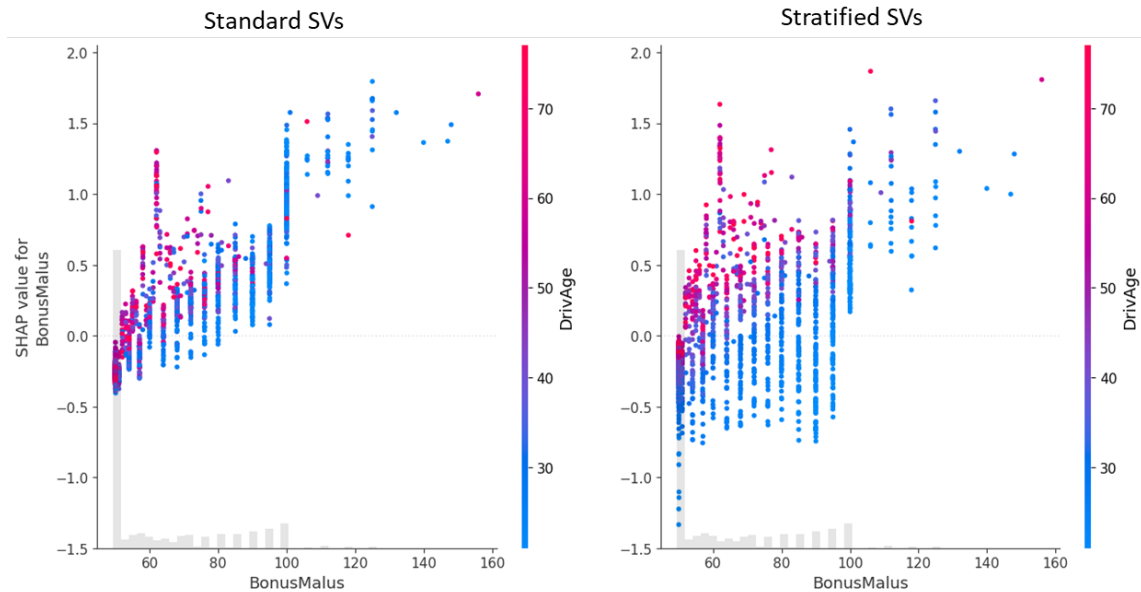

Figure 7: Shapley values for Bonus Malus feature for standard and stratified approaches.

## 5 Conclusions

In this paper we explored the impact of the correlated features on Shapley values calculated using marginal averaging. It was demonstrated on the example of a simple linear spline model that extrapolation caused by marginal averaging can lead to unintuitive results. Another undesirable effect is that the attributions would be model dependent as different types of extrapolations could produce different results.

To address the issues caused by model extrapolation a "stratified" approach was proposed where Shapley values are calculated for each stratum of the feature separately. While this has led to more intuitive attributions, the side effect was that each stratum had different reference value. We showed that a common reference can be obtained by utilizing causal information with an additional benefit of flexibility of providing explanations relative to different types of observations.

Remarkably, while stratified approach uses marginal averaging, it has led to the same results as causal Shapley values which rely on conditional averaging, at least for a simple problem considered here. This may point to a way of estimating the causal Shapley values for more complex problems without the use of conditional averaging.

The viability of the stratified approach when applied to real data was demonstrated on the example of French motor third party liability claims frequency data. Interestingly, stratified approach allowed us to discover interaction between driver's age and Bonus Malus score which wasn't as apparent in standard Shapley values.

## A   Shapley Values with Constant Extrapolation

To capture the constant extrapolation shown in Figure 1b, the Eq. (3.1) can be written as

$$f(X_1, 0) = \beta_0 + \beta_1 X_1 I(X_1 \leq 0)$$
$$f(X_1, 1) = \beta_0 + (\beta_1 + \beta_{12}) X_1 I(X_1 > 0) \quad \text{(A.1)}$$

Then we have

$$v(\{1\}) = E[f(x_1^*, X_2)] = f(x_1^*, 0) P(X_2 = 0) + f(x_1^*, 1) P(X_2 = 1)$$
$$= \frac{1}{2}(f(x_1^*, 0) + f(x_1^*, 1)) \quad \text{(A.2)}$$
$$= \beta_0 + \frac{\beta_1 x_1^*}{2} + \frac{\beta_{12}}{2} x_1^* I(x_1^* > 0)$$

The other averages are

$$E[f(X_1, 0)] = E[\beta_0 + \beta_1 X_1 I(X_1 \leq 0)]$$
$$= \beta_0 + \beta_1 E[X_1 I(X_1 \leq 0)] \quad \text{(A.3)}$$
$$= \beta_0 - \beta_1 \gamma$$

and

$$E[f(X_1, 1)] = \beta_0 + (\beta_1 + \beta_{12})\gamma \quad \text{(A.4)}$$

Putting everything together we have

$$v(\{2\}) = E[f(X_1, x_2^*)] = \beta_0 - 2\beta_1 \gamma (0.5 - x_2^*) + \beta_{12} \gamma x_2^* \quad \text{(A.5)}$$

The remaining value functions $v(\{\emptyset\})$ and $v(\{1,2\})$ are the same as in Eq. (3.2). Substituting value functions into Eq. (3.3), we obtain Shapley values for the linear spline model with constant extrapolation in Table 2.

## B   Causal Shapley Values

The formulas for causal Shapley values from [8] and [9] are shown in Table 5. As described in those works, the causal Shapley values are derived from the usual formulas in Eqs. (3.3) by only retaining the terms that satisfy assumed causality. Note, that conditional averaging is used for the valuation function i.e. $v(S) = E[f(X)|X_S = x_S^*]$.

Table 5: Causal Shapley values for the model with two features for two different causal relationships

|  | $X_1 \rightarrow X_2$ | $X_2 \rightarrow X_1$ |
|---|---|---|
| $\phi_0$ | $v(\{\emptyset\})$ | $v(\{\emptyset\})$ |
| $\phi_1$ | $v(\{1\}) - v(\{\emptyset\})$ | $v(\{2,1\}) - v(\{2\})$ |
| $\phi_2$ | $v(\{1,2\}) - v(\{1\})$ | $v(\{2\}) - v(\{\emptyset\})$ |

The value functions $v(\{\emptyset\})$ and $v(\{1,2\})$ are the same as in Eq. (3.2); therefore, we only need to evaluate $v(\{1\})$ and $v(\{2\})$. For $v(\{1\})$ we have

$$v(\{1\}) = \beta_0 + \beta_1 x_1^* + \beta_{12} x_1^* E[X_2 | X_1 = x_1^*] \tag{B.1}$$

with

$$E[X_2 | X_1 = x_1^*] = \frac{1 + sign(x_1^*)}{2} \tag{B.2}$$

For $v(\{2\})$ we have

$$v(\{2\}) = \beta_0 + (\beta_1 + \beta_{12} x_2^*) E[X_1 | X_2 = x_2^*] \tag{B.3}$$

with

$$E[X_1 | X_2 = x_2^*] = sign(x_2^* - 0.5) 2\gamma \tag{B.4}$$

Substituting value functions into equations in Table 5 we obtain causal Shapley values in Table 6.

Table 6: Causal Shapley values for linear spline model

|  |  | $x_1^* \leq 0, \quad x_2^* = 0$ | $x_1^* > 0, \quad x_2^* = 1$ |
|---|---|---|---|
| $X_1 \to X_2$ | $\phi_0$ | $\beta_0 + \beta_{12}\gamma$ | $\beta_0 + \beta_{12}\gamma$ |
|  | $\phi_1$ | $\beta_1 x_1^* - \beta_{12}\gamma$ | $(\beta_1 + \beta_{12})x_1^* - \beta_{12}\gamma$ |
|  | $\phi_2$ | 0 | 0 |
| $X_2 \to X_1$ | $\phi_0$ | $\beta_0 + \beta_{12}\gamma$ | $\beta_0 + \beta_{12}\gamma$ |
|  | $\phi_1$ | $\beta_1(x_1^* + 2\gamma)$ | $(\beta_1 + \beta_{12})(x_1^* - 2\gamma)$ |
|  | $\phi_2$ | $-(2\beta_1 + \beta_{12})\gamma$ | $(2\beta_1 + \beta_{12})\gamma$ |